\pdfoutput=1

\documentclass{acmart}

\usepackage[utf8]{inputenc}

\usepackage{csquotes}

\usepackage{url}
\usepackage{xcolor}
\usepackage{pdfpages}

\usepackage{gb4e} 
\noautomath 

\begin{document}

\title{Theories of ``Gender'' in NLP Bias Research}
\author{Hannah Devinney}
\affiliation{%
  \institution{Ume\aa{} University}
  \city{Ume\aa{}}
  \country{Sweden}}
\email{hannahd@cs.umu.se}

\author{Jenny Bj\"{o}rklund}
\affiliation{%
  \institution{Uppsala University}
  \city{Uppsala}
  \country{Sweden}}
\email{jenny.bjorklund@gender.uu.se}

\author{Henrik Bj\"{o}rklund}
\affiliation{%
  \institution{Ume\aa{} University}
  \city{Ume\aa{}}
  \country{Sweden}}
\email{henrikb@cs.umu.se}

\date{May 2022}
\acmISBN{}
\acmYear{2022}
\acmBooktitle{Proceedings of the ACM Conference on Fairness, Accountability and Transparency (ACM FAccT)}
\copyrightyear{2022}

\begin{abstract}
    The rise of concern around Natural Language Processing (NLP) technologies containing and perpetuating social biases has led to a rich and rapidly growing area of research. Gender bias is one of the central biases being analyzed, but to date there is no comprehensive analysis of how “gender” is theorized in the field. 
We survey nearly 200 articles concerning gender bias in NLP to discover how the field conceptualizes gender both explicitly (e.g. through definitions of terms) and implicitly (e.g. through how gender is operationalized in practice). In order to get a better idea of emerging trajectories of thought, we split these articles into two sections by time.

We find that the majority of the articles do not make their theorization of gender explicit, even if they clearly define “bias.” Almost none use a model of gender that is intersectional or inclusive of nonbinary genders; and many conflate sex characteristics, social gender, and linguistic gender in ways that disregard the existence and experience of trans, nonbinary, and intersex people. There is an increase between the two time-sections in statements acknowledging that gender is a complicated reality, however, very few articles manage to put this acknowledgment into practice.
In addition to analyzing these findings, we provide  specific recommendations to facilitate interdisciplinary work, and to incorporate theory and methodology from Gender Studies. Our hope is that this will produce more inclusive gender bias research in NLP.
\end{abstract}

\keywords{natural language processing, gender bias, gender studies}

\maketitle

\section{Introduction} \label{s:intro}

Algorithmic fairness and ``social bias''\footnote{Prejudice and/or stereotyping against particular social groups, which may result in direct or indirect discrimination.} are matters of increasing concern in the field of Natural Language Processing (NLP). The field's language models encode human prejudices and stereotypes including gender biases (see, e.g.,~\cite{Bolukbasi2016, Caliskan2017, Garg2018}). Although there are meta studies on how \textit{bias} is (or fails to be) theorized and operationalized, there is no corresponding work on how \textit{gender} is theorized and operationalized in NLP bias research. 
This article seeks to fill this gap: expanding on~\citeauthor{Cao2020}~\cite{Cao2020}, we survey 176 articles concerned with identifying and/or mitigating ``gender bias'' in NLP.

We question both how papers discuss and/or define ``gender,'' and how these definitions are implemented. In order to investigate this, we read the articles that make up our survey with the following research questions in mind: 
Is a theory of gender discussed, and if so, which one(s)? Do these theories draw from literature outside of NLP, such as Feminist, Gender, and Queer Studies? Where a theory of gender is not discussed, what does the underlying theory behind the method seem to be? How do gender theorization and operationalization connect to definitions and measures of ``bias''?

\section{Background} \label{s:background}

\subsection{Related Work} \label{s:related_work}

\citeauthor{Blodgett2020} \cite{Blodgett2020} survey articles analyzing and/or mitigating ``bias.'' They find that these articles tend to lack normative reasoning; specificity regarding what ``bias'' is and who it harms; and grounding in theories from outside mainstream NLP. Additionally, proposed methods ``are poorly matched to their motivations.'' 

Focusing specifically on gender bias in NLP, \citeauthor{Sun2019} \cite{Sun2019} categorize approaches to detecting and mitigating bias. They note limitations of these approaches, including that we have little idea how they behave at scale, because many focus on small parts of larger systems and are only verified for limited applications.
\citeauthor{Cao2020} \cite{Cao2020} study cisnormativity\footnote{The assumption that all people are cisgender, i.e. that their gender identity matches the sex they were assigned at birth.} in published NLP papers, focusing in particular on coreference resolution. They find that pronouns other than \textit{he} and \textit{she} are rarely considered, and social or personal gender is typically not distinguished from linguistic or grammatical gender.
\citeauthor{Savoldi2021} \cite{Savoldi2021} address how gender bias is conceptualized in machine translation. Their findings emphasize the need for understanding both the relationships between gender and language and the ways different factors can contribute to gender bias.  

All of these articles call for more interdisciplinarity. 
We echo this and provide some specific recommendations to facilitate such work in Section~\ref{s:collaboration}. 
We also discuss incorporating theory and methodology from the fields of Feminist, Gender, and Queer Studies into NLP research.

\subsection{Gender} \label{s:gender}

\paragraph{The ``Folk'' Model.} \label{s:folk_gender}
Discussing the operationalization of gender in Human Computer Interaction research,~\citeauthor{Keyes2018} ~\cite{Keyes2018} describes the ``folk understanding" of gender, where gender is derived from sex and the two are often conflated. In this model, gender is \textit{binary}, \textit{immutable}, and \textit{physiological}: there are two categories (man or woman), a person cannot alter their assigned category, and assignment is based on ``externally expressed physical characteristics"~\cite{Keyes2018}. This corresponds in some ways to \citeauthor{ButlerGenderTrouble}'s idea of a ``stable gender" in the ``heterosexual matrix" - sex entails gender entails desire towards the ``opposite" category \cite{ButlerGenderTrouble}.


This model is not accurate: ``sex'' is not binary (intersex people exist, see~\cite{FaustoSterling_2000}, among others); gender is neither immutable nor binary (trans and nonbinary people exist, and many cultures worldwide recognize more than two genders); and people do not actually assign gender to others based on physiology~\cite{Keyes2018}. 

The ``folk" model erases the existence of trans and nonbinary people. This erasure from what is ``acceptable'' or ``normal'' has material and often violent consequences for trans people \cite{UN_humanrights, ACLU_violence}, who are not seen as having an ``intelligible'' gender by the people and systems they encounter. 
It additionally presents problems for cis people, notably cis women, reducing them to a shallow stereotype which may be at odds with their lived experiences, and defining them by their bodies (a form of objectification). 
This model is at odds with ``fair'' NLP systems, because it is guaranteed to exclude vulnerable populations from its reasoning.

\paragraph{Gender Performativity.} \label{s:performance}
In Feminist and Gender Studies, gender is understood 
as a social construction that varies by culture. 
``Sex'' and ``gender'' are decoupled. One of the most prevalent ways of thinking about this decoupling is through \textit{gender performativity}~\cite{ButlerGenderTrouble}. 

Gender performativity means that we construct gender via discursive practices: gender is what one \emph{does} rather than what one \emph{is}. Repeated acts and interactions over time create our shared understandings of gendered categories, how we ourselves fit (or do not fit) into them, and how we categorize others.
This theory rejects a stable gender: both as something culturally constant (acts may be interpreted differently in different times and places) and as something that is a constant property of individuals (a person's performativity may change).

Language is a part of gender performativity, and -- importantly for bias research -- a key part of how we transmit and maintain stereotypes \citep{Maass1996}, (re)produce meaning \citep{HallWork13}, and navigate systems of power. 
On the level of interaction between individuals, language acts can be used to accomplish particular goals, known in pragmatics as speech act theory. These acts may communicate intent (``Please close the window.'') or actively change the state of the world (``I now pronounce you married.''). 
Language acts can be part of performing gender. For example, introducing yourself with the name and pronouns you would like to be called contributes to how your gender performativity is perceived~\citep{ConrodTalk2020}. 
Language data for NLP are typically large corpora, and in aggregate do not reflect \textit{individual} gender performances and experiences, but rather the production of gendered categories.

Although \citeauthor{ButlerGenderTrouble}'s gender performativity is not the only possible model for gender (see chapter four of~\citeauthor{ConnellandPearse}~\cite{ConnellandPearse} for examples), it is one we view as fruitful when doing gender bias research in NLP. The discussions and recommendations in this article are  guided by it. 

As gender is conceived differently in different contexts, we must also question whether gender can be considered as a variable independent of all others. Gender intersects with other power asymmetries, such as race, class, sexuality, and (dis)ability (see, e.g.,~\cite{Crenshaw1991, framing_intersectionality}), and we must account for the specific, intersectional locations of minoritized groups in  NLP bias work.

\section{Survey} \label{s:survey}

\subsection{Method} \label{s:survey_method}

We collected papers concerned with ``gender bias'' over two phases. The first phase, collected in June and July of 2020, comprised 126 papers. We collected these from existing bibliographies of surveyed papers provided in~\cite{Blodgett2020, Dinan2019}. These cover papers ``analyzing `bias' in NLP systems ... [restricted] to papers about written text only''~\cite{Blodgett2020}. Based on titles and abstracts, we selected papers where gender was relatively significant: either the specific focus or used as a running example for more general tasks, such as bias measurement or mitigation. This yielded 115 titles, which we assume concern \textit{bias in NLP} on some level based on their presence in~\citeauthor{Blodgett2020}~\cite{Blodgett2020}. An additional 11, which fit both the NLP-bias and  gender-focus criteria, were added from the bibliography of~\citeauthor{Dinan2019}~\cite{Dinan2019}. All of these papers are included in the first round of the survey. 

In the second phase, we collected an additional 90 papers in September of 2021 following the method used by~\citeauthor{Blodgett2020}~\cite{Blodgett2020}, by searching keywords ``gender,'' ``bias,'' and ``NLP''/``Natural Language Processing" over several databases (the ACL anthology\footnote{\url{https://aclanthology.org/}}, Google Scholar\footnote{\url{https://scholar.google.com/}}, and arXiv\footnote{\url{https://arxiv.org/}}) and filtering for those published in 2020 and 2021. Five of these were later rejected either as duplicates or because they did not actually address gender bias in NLP. We randomly sampled 50 of the remaining papers for the second round of the survey. The full list of articles is in Appendix A.

We read the articles with the goal of identifying how they theorize and operationalize \textit{gender} and \textit{gender bias}. We also categorized each paper according to language investigated; what NLP technology the paper concerns; and whether gender bias is the main focus or a use case.
The headings are summarized in Table \ref{t:categories}. 
Papers can belong to zero categories or multiple categories under some headings, while others expect yes/no answers. For the latter, additional notes were taken (e.g. where gender bias was \textit{not} the focus, we noted if gender is presented in combination with other biases, demonstrating a general-purpose technique, etc.). 

\begin{table*}[ht]
\caption{Categorization Schema for Surveyed Papers}
\centering
\label{t:categories}
\begin{tabular}{p{0.25\textwidth}|p{0.65\textwidth}}
\textbf{Heading} & What is of concern? \\ \hline
\textbf{Gender} & How is gender theorized? \\
\textbf{Gender Bias} & How is gender bias theorized and/or measured? \\
\textbf{Technology} & What is the technology of interest? \\
\textbf{Gender Proxies} & How is gender operationalized? \\
\textbf{Gender Focus} & Is gender bias the focus of the paper? \\
\textbf{Language} & What language(s) are investigated? \\
\textbf{Binary Problematized} & Does the paper acknowledge that gender is not a binary attribute?
\end{tabular}
\end{table*}

\begin{table*}[ht]
\centering
\caption{Gender. How is gender theorized across papers? Note that papers may be included in multiple categories, so counts do not sum to 126 (round 1) or 50 (round 2).}
\label{t:gender_counts} 
\begin{tabular}{l|l|r|r}
    \textbf{Gender} & Inclusion Criteria &Round 1 & Round 2 \\\hline
    \textit{binary} & considers only men and women & 118 & 41 \\
    \textit{essentialist} & makes specific claim about fundamental difference  & 12 & 1 \\
    \textit{neutral} & includes concept of `neutrality' (mixed groups or unknowns) & 7 & 6 \\
    \textit{nonbinary-inclusive} & thinks \& operationalizes beyond two genders & 1 & 4 \\
    \textit{social construct} & acknowledges gender varies/is a social construction & 2 & 4 \\    
    \textit{trans-inclusive} & thinks \& operationalizes beyond cis men and women & 1 & 7 \\
    \textit{undefined} & no framework & 7 & 2 \\
    \textit{underspecified} & does not explicitly/clearly define gender & 62 & 24 \\
\end{tabular}
\end{table*}

\begin{table*}[ht]
\centering
\caption{Gender Proxies. What data is consider representative of gender? Note that papers may use multiple strategies, so counts do not sum to 126 (round 1) or 50 (round 2).}
\label{t:proxy_counts}
\begin{tabular}{l|l|r|r}
\textbf{Gender Proxy} & Inclusion Criteria & Round 1 & Round 2 \\\hline
\textit{pronouns} & uses gendered pronouns e.g. \textit{he, she, ze, they} & 34 & 18 \\
\textit{annotation} & relies on annotated gender labels (any kind) & 12 & 6 \\
\textit{word lists} & compiles (unordered) lists of gender-associated terms & 31 & 15 \\
\textit{word pairs} & matches pairs of "equivalent but for gender" terms & 26 & 9 \\
\textit{sentence pairs} & like word pairs, but only use full sentences  & 2 & 0 \\
\textit{grammatical gender} & uses grammatical (not lexical) gender e.g. morphological markers & 6 & 4 \\
\textit{names} & uses first names & 21 & 8 \\
\textit{author gender} & round 1: inferred by annotators; round 2: self-identification & 5 & 1 \\ 
\textit{photo (human-label)} & uses profile photos (labeled by annotators) & 3 & 1 \\
\textit{photo (AGR)} & uses profile photots (labeled by automated gender recognition) & 3 & 0 \\
\textit{unspecified or N/A} & does not specify \textit{or} does not use a proxy & 11 & 2 \\
\end{tabular}
\end{table*}

\begin{table*}[ht]
\centering
\caption{Gender Bias. How is gender bias theorized and/or measured?  Note that papers may use multiple strategies, so counts do not sum to 126 (round 1) or 50 (round 2).}
\label{t:bias_counts}
\begin{tabular}{l|l|r|r}
\textbf{Gender Bias} & Inclusion Criteria & Round 1 & Round 2 \\\hline
\textit{activities} & measures relationships between verbs and gendered nouns  & 1 & 0 \\ %
\textit{allocation} & concerned with allocational parity (any kind) & 1 & 0 \\ %
\textit{associations} & tests associations by comparing between gendered categories & 26  & 16 \\ 
\textit{clustering} & word embeddings: do non-definitionally gendered words cluster? & 5 & 0 \\ 
\textit{counterfactual} & a decision should be independent from gender attributes & 4 & 2 \\ 
\textit{data imbalance} & the problem is in balance in number/type of datapoints  & 3 & 5 \\ 
\textit{demographic parity} & unequal representation in model/results & 2 & 9 \\
\textit{denigration} & addresses misrepresentation (denigration/hate speech) & 5 & 1\\ 
\textit{downstream task} & compares performance on a downstream task & 2 & 3 \\ 
\textit{erasure} & addresses lack of representation  & 1 & 1\\ 
\textit{gender unaware} & gender should not be predictable & 2 & 3\\ 
\textit{gender vector} & word embedding: identify "gender" vector \& locate words  & 17 & 7\\
\textit{multiple} & explicitly considers several dimensions of bias & 2 & 2\\ 
\textit{occupations} & uses "occupation" titles as a way to measure bias & 22 & 16 \\ 
\textit{performance parity} & correctness ratio, incorporates > 2 genders & 11 & 6\\ 
\textit{perforamnce parity (M/F)} & correctness ratio for women vs men & 51 & 21\\
\textit{real world distributions} & compare representation to 'real' statistics & 1 & 2\\
\textit{sentiment} & sentiment detection specifically & 2 & 0\\ 
\textit{stereotype} & addresses misrepresentation (stereotypes) & 5 & 15\\ %
\textit{translation accuracy} & asks: is this gendered translation is correct? & 4 & 5 \\ %
\textit{undefined} & does not define their metrics & 1 & 0\\ %
\textit{underspecified} & does not address what 'bias' or 'fairness' is & 1 & 0\\ %
\textit{WEAT} & uses WEAT (or SEAT) as a measure, specifically & 17 & 7\\ %
\textit{word-pair direction} & word embedding: difference in "equivalent" words & 2 & 0\\ 

\end{tabular}
\end{table*}

In our analysis of the results, we focus on the categories surrounding gender (how it is theorized and operationalized; and if and how it is problematized).

\subsection{Results} \label{s:survey_results}

\subsubsection{First Round (126 papers)}
Our findings with respect to \textbf{Technology} and \textbf{Gender Bias} are consistent with \citeauthor{Blodgett2020}'s analysis of the NLP tasks covered and bias categories, respectively. The majority of the papers we survey (82,5\%) deal only with English; seven papers (5,5\%) concern machine translation; four (3,1\%) are about a single non-English language; and the remainder compare multiple languages.

\paragraph{Gender.}
Of the 126 papers surveyed, 77 (61,1\%) are categorized as having \textbf{Gender Focus}, meaning they are specifically concerned with ``gender bias" rather than using gender as an example bias. 
45,2\% of the articles were tagged neither \emph{underspecified} nor \emph{undefined}. These discuss or define ``gender'' to some extent within the text, although in many cases rather shallowly and without citations. 
49,2\% were tagged \emph{underspecified}: these papers never outright discuss what ``gender'' is, and instead seem to take it for granted that the definition is obvious to the reader. A rough idea of their definition can be discerned by analyzing methods and results to see how gender is operationalized.
The remaining 5,5\% (7 papers) are tagged \emph{undefined}. This indicates that not only is gender never discussed or defined, but also that the model used cannot be inferred. These papers are not tagged \textit{binary} because by not (explicitly or implicitly) defining gender they do not meet the criterion for this tag, ``considers (only) men and women.'' 

In fact, regardless of whether or not gender is defined, the methods of most studies (118, or 93,6\%) operationalize gender with a binary model that reflects the ``folk understanding'' of gender. This figure includes all of the 62 papers that are tagged \emph{underspecified}. Most papers concerning English language technology use pronouns, first names, word pairs, lists of words, or some combination of these strategies to identify referent gender. 
Where the papers are concerned with author gender or image captioning, this information is generally labeled based on (binary) human annotation or automated gender recognition from profile photos, which means there is a very strong risk of misgendering~\cite{Keyes2018}.

Nearly every paper in this round fails to be inclusive of trans experiences, binary or nonbinary, in their methodology. 12 papers (9,5\%) are tagged \emph{essentialist}, indicating experimental design or analysis that, regardless of the intent of the authors, specifically \textit{excludes} trans experiences by conflating gender presentation and bodies and/or assuming some essential difference between `men' and `women' exists.
These binary operationalizations, and their consequences, are discussed further in Section~\ref{s:cisnormativity_in_survey}.

Some papers acknowledge the existence of nonbinary people 
but note that they are excluded by the methodology used (see Section~\ref{s:genderbinary_in_survey}). Finally, there is one paper,~\cite{Cao2020}, which is trans-inclusive in theorization and methodology. Two other papers~\cite{Sap2020, Kang2019} meet the criteria for the \textbf{trans inclusive} and/or \textbf{nonbinary inclusive} categories but do not explicitly extend this inclusivity to their analysis (see Section~\ref{s:inclusivity_in_survey}).

\subsubsection{Second Round (50 papers)}

We find some slight improvement with regards to gender inclusivity in both theory and practice over the time between the first and second rounds. Only 41 papers (82\%) are tagged ``binary,'' a noticeable improvement from the first round (93,6\%); and 9 papers (18\%) are actively inclusive of multiple genders at least in their theoretical positioning. Unfortunately, more than half of these inclusive papers face limitations in actually operationalizing more than two genders (for example, limited nonbinary representation in participants or data). 
We do not find any improvement in the number of ``underspecified'' papers, where gender is not explicitly theorized (48\%, compared to 49,2\%); although those papers that \emph{do} include more detail on their theoretical grounding.

There is a large and encouraging increase in papers dealing with languages other than English: 42\% (of which about a quarter deal with machine translation), which is more than twice as many as in the first round, although it should be noted this evidence is somewhat circumstantial. One problem that is often focused on in these papers is that methods developed to identify bias in a Western, English-language context do not perform as well in other contexts, for example due to grammatical gender or differing stereotypes. 

\section{Discussion} \label{s:survey_discussion}

Our findings are largely similar across both rounds of the survey. Thus, the remainder of our presentation draws examples from and discusses themes and patterns found throughout all 176 papers.

Generally, we find that papers concerned with ``gender bias'' look \emph{only} at gender, attempting to separate it out from other variables. Although papers concerned with ``bias'' in a general sense are more likely to analyze multiple sources of bias, these sources are almost always analyzed separately, i.e. not in an \emph{intersectional} manner. Papers that do attempt intersectional analysis primarily explore race and gender in a U.S. context.

Nearly every paper surveyed seems to make the assumption that gender is binary and immutable (similar to findings in~\cite{Keyes2018} and~\cite{Cao2020}). Social gender and linguistic\footnote{Both grammatical and lexical-semantic.} gender are also consistently conflated. Many papers do not ever define gender: the title and/or abstract notes that the subject of the paper is ``gender bias'' and the reader must conclude that ``gender'' is the taken-for-granted difference between two groups, implicitly cis men and cis women. This lack of clear definition does not change across the two rounds of the survey, even as there are some changes in how gender tends to be operationalized.

Although the differences between the first and second round of the survey suggest a trend towards increased inclusion of nonbinary language, as well as an increase in acknowledging the necessity of this inclusion, the field still overwhelmingly ignores non-normative genders. 
We discuss this, as well as some of the roadblocks faced by researchers trying to include nonbinary people and language in their work, in section \ref{s:genderbinary_in_survey}.

\subsection{Cisnormativity} \label{s:cisnormativity_in_survey}

We find that the prevalence of cisnormative assumptions constrains thinking about gender (as in~\cite{Keyes2018}). 
The assumption of binarity  becomes a standard: some authors explain that ``for gender in semantics, we follow the literature and address only binary gender'' ~\cite{Zhou2019}. This culture of internal citation, discussed in ~\cite{Blodgett2020}, reinforces the lack of engagement with the theory of gender even in research that is \textit{about} gender. 
More worryingly, the binarity becomes unchallengeable, something that \emph{must} be assumed in order to fit in with the established literature: ~\citeauthor{Dayanik2021} ``assume a binary gender classification (male/female) to be compatible with existing datasets'' and then immediately assert that this choice ``should not be understood as a rejection of non-binary gender''~\cite[p. 52]{Dayanik2021}. Regardless of the reasoning (the convenience of existing models, or perhaps an expectation in the field that all new datasets will resemble the old datasets), this remains a clear rejection of nonbinary genders and people. 

The assumption of some inherent difference between ``the" two genders, has been problematized, repeatedly, in many fields (e.g.~\cite{ButlerGenderTrouble,FausoSterling_1992,Hird2000}) but here NLP seems to fall behind.
Even in papers where features other than gender (such as race or religion) are considered to have multiple categories, gender is consistently taken as ``the binary case," often uncritically. Many measures to both calculate and mitigate gender bias rely on the binary, and may treat binarization as a limitation for other features, without considering it to be one for gender.
The binary can be seen in common methodological choices (section \ref{s:groups_indiv}), such as collecting ``pairs" of words, the meanings of which differ only in gender.
Masculine and feminine genders are also generally presented as ``opposites" which can be ``swapped" for each other; placed on each end of a linear scale of bias; or used to define  vector space directions.

Although binary gender models \textit{can} be inclusive of binary trans experiences of gender, this is not often the case in the papers we surveyed. Gender is frequently tied to bodies, with language such as ``males and females'' (see  section~\ref{s:inclusive_lang}). The implicit assumption that everyone is cisgender is evident in both methods and analysis. 
This exclusion of nonbinary genders in analyzing language technology can contribute  to harms against nonbinary people, such as erasure; misgendering; derogatory associations; and allocative harms such as automatically rejecting applications to jobs or public services \cite{Dev2021}.

\paragraph{Cisnormative Methodology.} \label{s:cisnormative_methods}
Cisnormative assumptions in methodologies often determine both research questions and methods, including the proxies chosen to identify gendered categories or to gender individuals (section~\ref{s:groups_indiv}). 
Leaving out nonbinary people and genders is the most common, but not the only, way that cisnormativity is evident in most methods surveyed.

Methods that would otherwise be inclusive of binary trans people and experiences are rendered trans-exclusive by the choice of word lists or word pairs used. The word lists created by~\citeauthor{Zhao2018b}, which consist of ``words associated  with gender by definition''~\cite{Zhao2018b} are used in several papers 
to calculate gender in word embeddings. The two lists include terms like \textit{uterus, penis, testosterone} and \textit{ovarian cancer}. Although not based on an individual's appearance, these words incorporate the physiological assumption of the folk model. 

Another problem is assuming these physiological words can be ``paired'' in the same way that \textit{man:woman} or \textit{king:queen} can be. For example, the analogy ``\textit{she} to \textit{ovarian cancer} is as \textit{he} to \textit{prostate cancer}'' is categorized as ``gender appropriate''~\cite{Bolukbasi2016}, which erases those for whom these body parts and pronouns do not `match.'
These are not equivalent diseases. Ovaries and prostates are not ``equivalent'' body parts (a questionable concept) differing only by ``sex''.


\paragraph{Cisnormative Analysis.} \label{s:cisnormative_analysis}
Independent of method, the analysis of results can erase the lived reality of many trans people, e.g., by defining sentences such as \textit{he gave birth} as ``meaningless''~\cite{ZhangG2020},  ``nonsensical''~\cite{Sun2019}, or even ``biologically \ldots inaccurate facts''~\cite{Kocijan2020}. 
\textit{He gave birth} is not a statistically likely sentence in English, but it is neither meaningless nor nonsensical. It is certainly not ``biologically inaccurate'' -- trans men, and other people who use he/him but were assigned female at birth, can and do give birth in increasing numbers~\cite{Riggs2020}.

The idea of semantic incorrectness \textit{must} be understood in context, and it is necessary to draw a distinction between ``flipping'' gendered words that refer to \emph{specific people}, words that refer to \emph{unknown persons}, and words that refer to \emph{groups}. Consider the argument for counterfactual data substitution given in (\ref{Lu18_ex}):
\begin{exe}
    \ex\label{Lu18_ex} Flipping a gendered word when it refers to a proper noun such as \textit{Queen Elizabeth} would result in semantically incorrect sentences. ~\cite{Lu2018} 
\end{exe}
True, it would be misgendering to use \textit{he} in reference to the real person Queen Elizabeth II of England, or to call her a \textit{king}. 
However, it is not universally true to say that it is ``semantically incorrect'' for anyone named Elizabeth to co-refer with \textit{king} and/or \textit{he} (or \textit{monarch}, \textit{they}, or \textit{ze}). Coreference in English depends on context, both at the level of a particular conversation and at the level of world knowledge \cite{Ackerman2019}. Although names are associated with gender, they do not exclusively correspond to particular pronouns, and a system that assumes they do will ultimately commit errors including misgendering. 

Gendered pronouns and nouns do not automatically correspond to bodies, nor how we dress and adorn those bodies (aspects of gender presentation), as implied in (\ref{Kaneko19_ex}): 
\begin{exe}
    \ex\label{Kaneko19_ex} \ldots one would expect `beard' to be associated with male nouns and `bikini' to be associated with female nouns, and preserving such gender biases would be useful \ldots for a recommendation system  ~\cite{Kaneko2019}
\end{exe}
There may be a statistically strong \emph{association} between masculine nouns and beards, and beards are often a part of masculine performativity, but that does not make it a foolproof indicator of gender for an individual. Analyzing it as such erases trans and gender non-conforming people, and cis women with facial hair. Preserving this association in recommendation systems can be psychologically harmful for trans people who may be misgendered by ads. 
The particular example of \textit{beard} with ``male nouns'' is also ironic, as \textit{beard} can in specifically refer to a woman whom a gay man is dating to hide his sexuality -- making it a feminine noun in these cases. 

Factoring in this context is essential for analyzing NLP in a way that is accepting of gender diversity. We must know both what world (Queen Elizabeth of England, or King Elizabeth of the \textit{Pirates of the Caribbean} franchise?) and what discourse (the beard on someone's face, or the beard they are dating?) make up the context to make a judgement about how meaningful a particular sentence or association is.

\subsection{Gender Proxies} \label{s:groups_indiv}

One important design choice in any system for measuring or reducing gender bias is the strategy used to find evidence of gender, i.e. what information will be used as a proxy for gender. Papers concerned with ``gender bias'' in NLP are, broadly speaking, either concerned with bias based on who \textit{authored} the text, or based on who is \textit{referred to} in the text. Within both of these categories, asking the right questions about power and bias requires knowing whether to look at the gender of \textit{individuals} or gendered \textit{groups}.

\paragraph{Individuals.}
For cases where a system may be biased against authors of a particular gender, it is imperative to keep in mind that each of these authors is a  human with agency -- including over their gender identity and expression. \citeauthor{Larson2017} provides four key guidelines for the responsible use of (author) gender as a variable in NLP, namely:
\textit{i}) make theory of gender explicit; \textit{ii}) avoid using gender unless necessary; \textit{iii}) make category assignment explicit; and \textit{iv}) respect persons~\cite{Larson2017}.

We discuss \textit{i} in Section~\ref{s:clear_theory}. Clearly in research regarding gender bias it can be necessary to use gender as a variable, covering \textit{ii}. 
Therefore we primarily discuss proxy choice for individuals with respect to \textit{iii} and \textit{iv}. \citeauthor{Larson2017} makes it very clear that, where possible ``participant self-identification should be the gold standard for ascribing gender categories'' \cite[p. 7]{Larson2017} and discusses some of the specific challenges for obtaining this self-identification (e.g. the options do not adequately describe the participant's gender, or someone may choose not to provide their gender on platforms where it is not required). In cases where self-identification is not possible, the procedure for assigning participants to gendered categories must both be clear, reproducible, and respectful.

The most common proxies for assigning gender to an individual author are \emph{first names}, \emph{profile photos}, or a combination. In most cases in the papers we survey, automated methods such as facial recognition or gendering of first names based on statistics 
are used. These automated methods do make category assignment in theory reproducible, but fail the fourth requirement: ``respect persons''.
The use of facial recognition for automated gender ``recognition'' (i.e. assignment) is trans-exclusive:  ``essentialising the body as the source of gender'' \cite[p. 11]{Keyes2018} and typically representing gender as a binary attribute. Additionally, it has been shown that commercial gender classification systems are systematically worse for people of color, in particular for dark-skinned women \cite{Buolamwini2018}. Many people also do not use photos of their own face as a profile photo for a variety of reasons.

Assigning gender by first name  has particular pitfalls at the level of the individual. Among the surveyed papers, all strategies for using this proxy assume that gender is binary. Thus, nonbinary people will be consistently misgendered.

These strategies assume that a name has a ``gender'' if  a majority of the population with that name belongs\footnote{By sex assigned at birth, or legal gender: methods for assigning ``belonging'' vary.} to that gender category. 
There is a particular cultural bias in this assumption: while common for many Western cultures, reliably gendered names are not a universal constant. The gendered association of a particular name can change over time or by culture/language (e.g. ``Jean'' in French vs English). Limiting a study to names we can ``confidently'' gender risks skewing results towards the cultural majority for whom this data is available. 

\paragraph{Groups.}
At a group level, first names have some more concrete advantages: it is possible to use them in intersectional analysis (by choosing names that are strongly associated with e.g. Black women vs white women in a certain generation in the U.S.), and in aggregate use they do not have the exact same risks of misgendering individuals. However, this method remains unreliable: not all languages and cultures have strongly gendered names, and picking only strongly gendered names therefore likely favors certain demographics. 

Word pairs and lists (see Section \ref{s:cisnormativity_in_survey}) occasionally also use limited first names. First names are not gendered by definition and are positioned in particular generations, races, and religions. As a proxy, they must be used with many other names. 

Word pair strategies are typically binary and the ``semantic equivalence'' of many pairs can be called into question. Consider \textit{bachelor:spinster} -- \textit{spinster} is pejorative while \textit{bachelor} is not; and there is no such thing as a \textit{spinster's degree}.

As most of the papers deal only with English, grammatical gender was significantly less common and generally was not used as a proxy for social gender.\footnote{For a discussion on the division(s) between grammatical and social forms of gender, see~\cite{Ackerman2019}.} Where grammatical gender was a factor, articles dealt with alignments of grammatical and social gender (e.g. in generating accurate machine translation results) or developing methods to calculate gender bias in languages with grammatical gender. 

In languages like English, where gender is relatively reliably marked in third person pronouns (\textit{he, she, they, ze,} etc.), pronouns are one of the better proxies for gender, particularly when more than \textit{he} and \textit{she} are used. These must still come with a caveat: they are \textit{associated} with particular genders, but not every referent of a given pronoun belongs to the same gender category. 

\subsection{Binary Gender as a Limitation} \label{s:genderbinary_in_survey}

Around one fifth of the papers surveyed in the first round include some textual acknowledgement that gender is not binary, but do not extend their methods to reflect this definition. This rate increases to about a third of the papers in the second round. 
These acknowledgements are typically limited to a sentence or two, and are usually included in the limitations section or formatted as footnotes. 
Disconnects between theory of gender (where specified) and the method chosen can give the impression that inclusivity was an afterthought rather than a limitation. For example~\citeauthor{Sun2019} note that ``Non-binary genders \ldots should be considered in future work'', but define gender bias as ``the preference or prejudice toward one gender over \emph{the other}'' (emphasis added) ~\cite{Sun2019}.

It is relatively common to consider inclusion of genders outside the binary to be a problem for the future. 
This should ideally be accompanied by a specific commitment to participation in doing the necessary work, such as ``we plan to extend \ldots genders (e.g. agender, androgyne, trans, queer, etc.)''~\cite{Kiritchenko2018}.

A coherent methodology aligns theory and method, or indicates where the theory cannot fully be operationalized, for example due to limitations of the technique, as in (\ref{Chang19_ex}), or the data, e.g.~\cite{Swinger2019} where ``only binary gender statistics'' are available. 
However, unspecified ``technical limitations'' may give the impression that the researchers did not attempt to develop inclusive methods.
\begin{exe}
    \ex\label{Chang19_ex} Our method unfortunately could not take into account non-binary gender identities, as it relied on she/her and he/his pronouns, and could not easily integrate the singular they/them, nor could we find sufficient examples of ze/zir or other non-binary pronouns in our data.~\cite{Chang2019}
\end{exe}

Encouragingly, there seems to be an increase in articles where the authors attempt to work around these limitations, one step at a time. \citeauthor{Yeo2020}~\cite{Yeo2020} and~\citeauthor{Vig2020}~\cite{Vig2020}, for example, both incorporate experiments involving singular \emph{they} but find several confounding factors (including the tendency for \emph{they} to be processed as plural) that prevent these results from being analyzed in the same way as their binary experiments. \citeauthor{Ramesh2021} provide extensive bias- and gender- statements, and attempt to adapt existing metrics to include more than just masculine and feminine; but are somewhat limited by the binary nature of many of these metrics~\cite{Ramesh2021}.
Combined with the observed increase in trans-inclusive methodologies, further discussed in the following section, it is clear that there is a desire to include minoritized groups despite how this may complicate research methods.


\subsection{Trans-Inclusive Methodology} \label{s:inclusivity_in_survey}

In their work on gender bias in coreference resolution, \citeauthor{Cao2020} present an encouraging example of what gender-inclusive NLP can look like~\cite{Cao2020}. A detailed theoretical grounding on the nature of both ``gender'' (in society and in language) and gendered harms (including trans-specific harms stemming from exclusion) motivates and supports the paper. They focus on the data side of coreference resolution tasks, and a key aspect of their method is how they source data: both obtaining permission from authors and working with stakeholder communities to develop a dataset ``by and about trans people''. 

Two papers have at least one annotator identifying themself as nonbinary or ``other''~\cite{Sap2020, Kang2019}. These papers meet the criteria for the \emph{trans} and/or \emph{nonbinary inclusive} tags, but do not necessarily analyze gender or their results in ways that account for this gender diversity. \citeauthor{Sap2020}'s annotation method leave spaces for trans and gender nonconforming issues, via free-text responses~\cite{Sap2020}, but this relies on annotators being familiar with trans people, issues, and stereotypes.

Papers in the second round (published after the spring of 2020) increasingly feature trans-inclusive methodologies. \citeauthor{Hansson2021}~\cite{Hansson2021} present a Wino-gender style for Swedish which incorporates the neutral third-person singular neo-pronoun \textit{hen}. \citeauthor{Dinan2020}~\cite{Dinan2020}  offer a framework for categorizing gender in dialogs along multiple dimensions, classifying genders as \{masculine, feminine, neutral, unknown\}. \citeauthor{Munro2020}~\cite{Munro2020} address the performance gap between \emph{his} and both \emph{hers} and \emph{theirs} in part of speech taggers.
In addition to papers included by random sample, we are aware of several other papers with trans- and nonbinary-inclusive methodologies.

\section{Recommendations} \label{s:recommendations}

Although it will never be possible to perfectly model the world in all its complexity, simplification of identity aspects such as gender willfully ignores the existence and experience of marginalized groups and can contribute to material harms done to already vulnerable populations. As researchers and practitioners concerned with the risks evident in biased NLP tools, we must be attentive to these harms and incorporate that knowledge into our work.

To address this, we make a series of recommendations for addressing theories of gender, choosing bias measures, writing about gender in a respectful manner, and incorporating feminist research methodologies.
In addition to these recommendations, we provide the start of an Open Bibliography, \textit{Gender Theory for Computer Scientists}\footnote{\url{https://github.com/hdevinney/open-gender-bib}}. This is a community resource for collecting readings and other reference material we find useful for addressing questions of gender applied to NLP.

\subsection{Make Theorization of Gender Explicit} \label{s:clear_theory}


First, we extend \citeauthor{Larson2017}'s first guideline for using gender as a variable (``make theory of gender explicit'') to all work on gender bias in NLP. The way that the operationalization of gender is grounded in this theory should also be specifically discussed.
This could be considered to already be a part of \citeauthor{Blodgett2020}'s call for explicit definitions of bias, which includes both how particular systems are harmful and \textit{to whom}. However, it is not enough to, for example, claim a particular behavior is harmful to women without some understanding of who this encompasses, because ``women'' may not be a coherent category (see, among others, ~\cite{ButlerGenderTrouble, hooks_feminism}). 
Grounding the conceptualization of gender in theory can also help specify the scope of the work and allows for better comparisons between studies. 

At the Second Workshop on Gender Bias in Natural Language Processing, a \textit{bias statement} was required for all submissions~\cite{gebnlp-2020-gender}. Authors were asked to clarify what harms, against whom, their research was concerned with. These statements were evaluated by reviewers from the Humanities and Social Sciences. In addition to providing an opportunity for reflexive analysis on the parts of the authors, these statements make normative assumptions explicit to reviewers, helping them evaluate the research in context.
For example, the bias statement in \citeauthor{Falenska2021}~\cite{Falenska2021} clearly defines the way that the `bias' in question harms people, and how this connects to their lived experiences, and makes reference to relevant literature outside of NLP. They explain exactly why they have limited their work to binary gender, but continue to revisit the question of nonbinary inclusion throughout the paper.

Making normative assumptions explicit is helpful both for reviewers and readers to evaluate and make sense of research, and for researchers themselves to formulate the most appropriate research questions and methods for their work. 
The latter works best when the requirement becomes part of the culture of research, rather than a checkbox when submitting a paper. When the model of gender used is explicitly defined from the start, researchers can ``check back'' as they develop their methods to ensure alignment (see Section~\ref{s:positionality}). For example, from the research side, when exploring gendered associations in news corpora for \citeauthor{DevinneyGBNLP}~\cite{DevinneyGBNLP}, we realized that despite our attempts to analyze gendered categories, our dataset contained little to no nonbinary representation. This resulted in the ``gendered, but not men or women'' category being primarily about mixed-gender groups. Because we checked back and realized early in the process that our data and methods did not align, we were able to gather more text in a different corpus which contained sufficient nonbinary representation to allow for our desired analysis.

\subsection{Understand Your Goal} \label{s:fairness}

The concept of ``bias'' is a complicated issue rooted in society and culture, and in order to make progress in correcting for unfairness in a technical domain it is essential that we understand what we are trying to accomplish.
Just as bias can be defined in many conflicting ways \cite{Blodgett2020}, so can its absence. Technology attempting to reach this state is often referred to as ``ethical'' or ``fair,'' but definitions vary.
From a data science perspective, \citeauthor{DataFeminism}~\cite{DataFeminism} divide concepts around removing ``bias'' into two types. Those that \textit{secure power} provide some short term solutions, but ultimately do not go far enough. Those that \textit{challenge power}, however, ``acknowledge structural power differentials and work towards dismantling them''~\cite[p. 60]{DataFeminism}. 

There are a number of measures of algorithmic fairness, including \emph{fairness through unawareness}, \emph{fairness through awareness}, \emph{demographic parity}, and \emph{counterfactual fairness}. They can broadly be classified as measures that work on a group level and measures that work on an individual level. (For a summary, see, e.g.,~\cite{Mehrabi2019ASO}.) There are also domain-specific measures of, e.g., how biased a particular word embedding is.
In order to determine what measure is best suited for a particular project, we require clear goals. What harms are we trying to prevent and against whom? If we are trying to ensure fair representation, group level measures seem appropriate; whereas if we want to avoid unfair hiring practices in a system, an individual level measure might be better suited.


Bias in language is a complicated topic, and we must not overstate the efficacy of mitigation measures. Claims to ``significantly reduce'' or ``eliminate'' gender bias must acknowledge that this can only be known for the measures of bias tested. Claiming otherwise overreaches the scope of our research and misrepresents the field of algorithmic bias research as something that \textit{has} a purely technical solution~\cite{Green2019}.
The fact remains that we (as researchers, and as a global society) do not know what a truly ``unbiased'' system is. Incremental work towards an unknown goal is difficult, so it can be useful to ask ourselves open questions, such as: In an unbiased world, what would an unbiased system look like? In our biased world, what does an unbiased system look like? Is the goal of NLP to understand and use language as humans \textit{currently} do? Are there things we would like NLP to be better than us at? 

\subsection{Use Consistent, Respectful, and Accurate Language} \label{s:inclusive_lang}


One challenge we faced in conducting this survey was that, in addition to individual papers failing to define their theorization of gender, there is no standard vocabulary within the field of ``gender bias in NLP''. This makes comparing papers ``within'' this field difficult, but also presents difficulties for connecting to ``outside'' research. The vocabulary chosen to describe different gender categories, aspects of grammatical gender, and individual people is not consistent with related literature from the fields of linguistics and gender studies. 
It is necessary to decouple the various sociolinguistic aspects of gender, i.e. ``the  different ways in which gender can be realized linguistically''~\cite{Cao2020}, in order to develop coherent methodologies and accurately analyze results.

In particular, we note the problematic use of the terms \textit{male} and \textit{female} in two circumstances: 1) as nouns describing people and 2) as adjectives describing grammatical or lexical gender.
Although it is not grammatically incorrect in English to use these terms as nouns, the implications and context of this usage are important because our work purports to try to counter gender biases in language.
Such language is regularly used to dehumanize vulnerable groups, including women, trans people of all genders, and people of color. 

As adjectives describing grammatical or lexical gender, this is simply not the preferred terminology in linguistics.
For languages with grammatical gender (i.e. noun classes defined by syntactic agreement) it may be traditional to refer to \textit{feminine}, \textit{masculine}, \textit{neuter} (etc.) words; in other languages these classes are numbered \cite{wals-30}. 

We thus consider \textit{male} and \textit{female} to be inappropriate terms within NLP and suggest instead that we follow  linguistics praxis when discussing grammatical and lexical gender,  and gender studies  when discussing people. This encompasses adjectives such as \textit{masculine, feminine, genderless, nonbinary, unknown, neutral} and nouns such as \textit{men, women, nonbinary people}. Where individual terms (e.g. pronouns) are investigated, the term can simply be specified without categorizing.


\subsection{Use Feminist Research Methodologies} \label{s:interdisciplinary}

In addition to incorporating theories of gender from other fields, we recommend borrowing and adapting methodologies from Feminist research, including reflexive research and situated knowledges. When the model of gender used is explicitly defined from the start, researchers can ``check back'' as they develop their methods to ensure alignment. 
We also recommend collaboration between researchers and other stakeholders to open up new avenues of inquiry. 

\paragraph{Situate Knowledge.} \label{s:positionality}
NLP research has a tendency to use what~\citeauthor{HarawayKnowledges} describes as the ``god trick''~\cite{HarawayKnowledges}.
By aggregating enough language data, the logic goes, we can actually achieve the ``view from nowhere'' and thereby produce objective truth. But knowledge does not come from nowhere, it is produced by people who are \textit{situated} in particular locations, with different backgrounds, experiences, viewpoints, etc. These positions inform how we think and what we notice, and are always partial and subjective. By interrogating the world from our various positions, we produce partial and situated knowledges, which can be compared and combined to triangulate something that approaches ``objective'' truth. We can also reflexively analyze our research, to better understand how our research methods and goals align.

%
Positionality statements are one way to situate ourselves and contextualize our research, problems, data, and solutions. 
They are tools which allow us the opportunity to reflect on what we see (and what we might miss); assumptions we may be making; and potential gaps in our knowledge. 
Research lacking in reflexivity is ``likely to reproduce the exact forms of social oppression'' we are trying to prevent~\cite[p. 2]{Green2019}.

Like explicit descriptions of theoretical groundings, when shared they can help the audience interpret how and why the author(s) developed their methods and reached their conclusions.
For example, with respect to this paper: all three authors of this paper are white academics based in Sweden, with EU citizenship. This positionality influences how we think about, e.g., race or immigration. We can read and learn from others but we cannot directly know or access their experiences.
Our academic backgrounds (in computer science, linguistics, and gender studies) and personal experiences both inform what we are likely to notice when analyzing ``gender'' in NLP. As a queer, gender non-conforming person, the first author is more likely to notice cisnormativity due to the extent that they confront and navigate it in their everyday life.

Although this reflexive examination of our positions and work should be a component of our research processes, it is not necessary to include it in reporting on that research. 
Particularly for minoritized researchers in a field where positionality statements are not an unremarked standard, it can be dangerous to make ones status as part of a minoritized group public. Therefore, it is \emph{not} our recommendation that such statements be disclosed publicly; rather, we encourage researchers to consider their own positionalities at every step of the research process, and reflexively interrogate their research in light of these positionalities.

\paragraph{Work with Others.} \label{s:collaboration}
One way to access other positions and knowledges is collaboration: with researchers across disciplinary lines and with stakeholders outside of academia and industry. Stakeholders offer lived expertise and experience and minoritized groups (such as queer and trans people) in particular must be included in NLP research that concerns us. The phrase ``nothing about us without us'', originating in disability rights activism, applies. If we are the ones at risk, we must have a hand in shaping the solutions: beyond surface-level inclusion which fails to challenge the social order~\cite{Hoffmann2020}. One place this applies directly is in collecting more data to compensate for sparsity, which must be done in collaboration with communities and with respect for autonomy. Being made visible and countable within the data, may both benefit minoritized groups and put them at risk.

Doing  inter-disciplinary work requires time and deliberation. It specifically strives to ``develop a joint understanding of the problem'' and melds approaches from the researchers involved~\cite[p. 4]{Mobjork2019}. 
Thus, much of this time is best spent explaining and listening, which allows all participants to make connections and figure out how best to blend them and reconcile any fundamental differences of opinion or understanding. 
We have found in our collaborative work that (global health situations allowing) sitting down together for longer periods of time leads to more progress, as this gives us opportunities to ask questions and make sure we understand the answers. 

\section{Conclusion} \label{s:conclusion}

We surveyed nearly 200 papers concerned with gender bias in NLP. 
Throughout both time-spans surveyed, the vast majority of papers do not discuss how they theorize gender and/or operationalize gender following the cisnormative ``folk model''. However, differences between the two rounds suggest that there is more and more research inclusive of nonbinary genders, as well as an increase in papers noting that the binary model of gender is a methodological limitation. 
In analyzing some of the problems with these papers, we detail how they exclude trans people and experiences and the impact this has on NLP bias research as a whole. 

Finally, we provide some recommendations for doing better gender bias research in NLP, such as explicitly defining gender (and using respectful language to do so); selecting methods that work well with this definition; and borrowing from feminist research methodologies. To help with the last aspect, we contribute the start of a community-sourced list of recommended texts and other resources concerning gender, language, and feminist theory.


\vspace{2ex}

\noindent\emph{Funding/Support.} The authors declare no additional sources of funding.

\bibliographystyle{ACM-Reference-Format}
\bibliography{survey_refs.bib, theory_refs.bib}

\includepdf[pages=-]{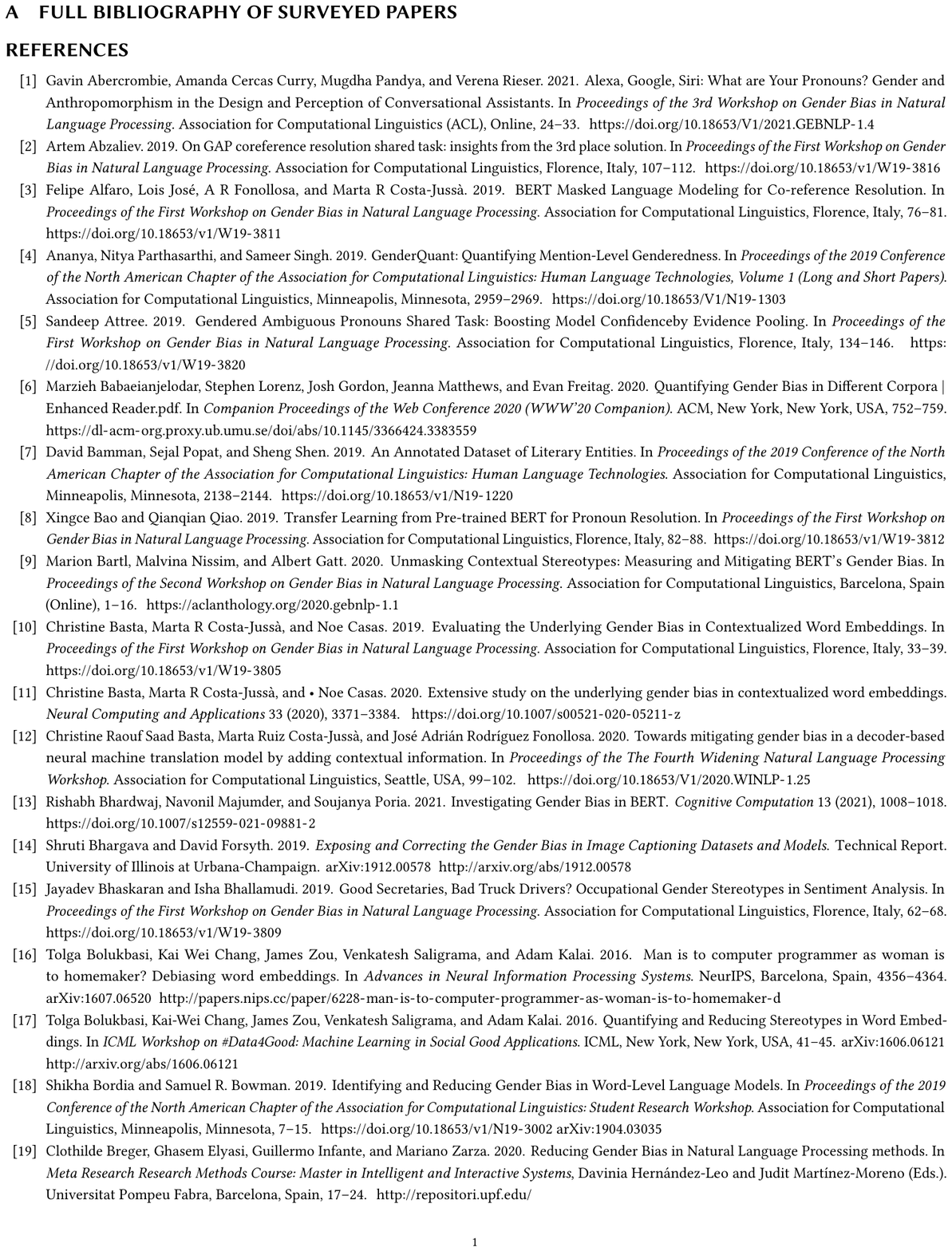}

\end{document}